\pdfoutput=1

\documentclass[11pt]{article}

\usepackage[]{acl}
\usepackage{stfloats}
\usepackage{times}
\usepackage{colortbl} 
\usepackage{latexsym}
\usepackage{hyperref}
\usepackage{graphicx}
\usepackage{mdframed}
\usepackage{xcolor}
\usepackage{multirow}
\usepackage{amsmath}
\usepackage{amssymb}
\usepackage{pifont}
\usepackage[final]{changes}
\definecolor{lightblue}{HTML}{E8F0FE}
\usepackage[T1]{fontenc}
\usepackage{float}
\usepackage[utf8]{inputenc}
\usepackage{adjustbox}
\usepackage{microtype}
\usepackage{array}
\usepackage{booktabs}
\usepackage{placeins}
\usepackage{ulem}
\usepackage{inconsolata}
\usepackage{tabularx}

\newcommand{\benchname}{CorrelationQA}
\newcolumntype{M}[1]{>{\centering\arraybackslash}m{#1}}

\NewDocumentCommand{\tyh}
{ mO{} }{\textcolor{blue}{\textsuperscript{\textit{tyh}}\textsf{\textbf{\small[#1]}}}}
\title{The Instinctive Bias: Spurious Images lead to Illusion in MLLMs}

\author{Tianyang Han$^{3}$\thanks{\, Equal Contribution. 
}, Qing Lian$^1$\footnotemark[1], Rui Pan$^2$\footnotemark[1], 
\textbf{Renjie Pi$^1$}, \\
\textbf{Jipeng Zhang}$^1$,
\textbf{Shizhe Diao}$^4$,
\textbf{Yong Lin}$^1$,
\textbf{Tong Zhang$^{2}$}
\\
$^1$The Hong Kong University of Science and Technology \\ \quad $^2$University of Illinois at Urbana-Champaign  \\
$^3$The Hong Kong Polytechnic University \quad $^4$NVIDIA \\
\\
}

\begin{document}
\maketitle

\begin{abstract}
Large language models (LLMs) have recently experienced remarkable progress, where the advent of multi-modal large language models (MLLMs) has endowed LLMs with visual capabilities, leading to impressive performances in various multi-modal tasks. 
However, those powerful MLLMs such as GPT-4V still fail spectacularly when presented with certain image and text inputs. 
In this paper, we identify a typical class of inputs that baffles MLLMs, which consist of images that are highly relevant but inconsistent with answers, causing MLLMs to suffer from visual illusion.
To quantify the effect, we propose \benchname{}, the first benchmark that assesses the visual illusion level given spurious images. This benchmark contains 7,308 text-image pairs across 13 categories.
Based on the proposed \benchname{}, we conduct a thorough analysis on 9 mainstream MLLMs, illustrating that they universally suffer from this instinctive bias to varying degrees.
We hope that our curated benchmark and evaluation results aid in better assessments of the MLLMs' robustness in the presence of misleading images. 
The code and datasets are available at \href{https://github.com/MasaiahHan/CorrelationQA}{https://github.com/MasaiahHan/CorrelationQA}. 
\end{abstract}

\section{Introduction} 
Large language models (LLMs) have sparked a transformative shift in the field of artificial intelligence~\cite{zhao2023survey, workshop2022bloom,chowdhery2023palm,touvron2023llama}. 
Following the development of LLMs, a series of multi-modal large language models (MLLMs) have emerged to enable LLMs with visual processing capabilities~\cite{alayrac2022flamingo,gong2023multimodal,yin2023survey,zhu2023minigpt}.
Typically, current MLLMs process visual inputs by converting them into visual tokens that share the same latent space as language tokens in LLMs. This conversion not only maintains excellent text processing abilities but also enables LLMs with powerful visual semantic understanding capabilities.
These models have demonstrated commendable performance in downstream tasks such as image captioning~\cite{hossain2019comprehensive, mplug} and visual question-answering~(VQA)~\cite{goyal2017making, pali}.
\begin{figure}[t]
    \centering
    \includegraphics[width=\columnwidth]{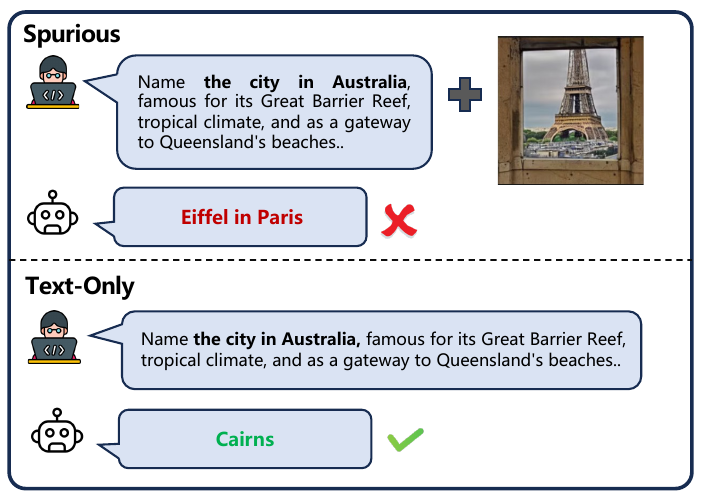}
    \caption{Cases of Instinctive Bias in LLaVA. \\
    \textbf{Top (Spurious image)}: when presented with images that are related but do not correspond to the correct answer (\textit{i.e.} Eiffel in Paris), MLLMs are hallucinated to provide an incorrect answer.\\
    \textbf{Bottom (Text-only)}: without spurious images, MLLMs display the ability to provide the correct answer.}
    \label{examplecase}
    \vspace{-6mm}
\end{figure}

\begin{figure}[t]
    \centering
    \includegraphics[width=\columnwidth]{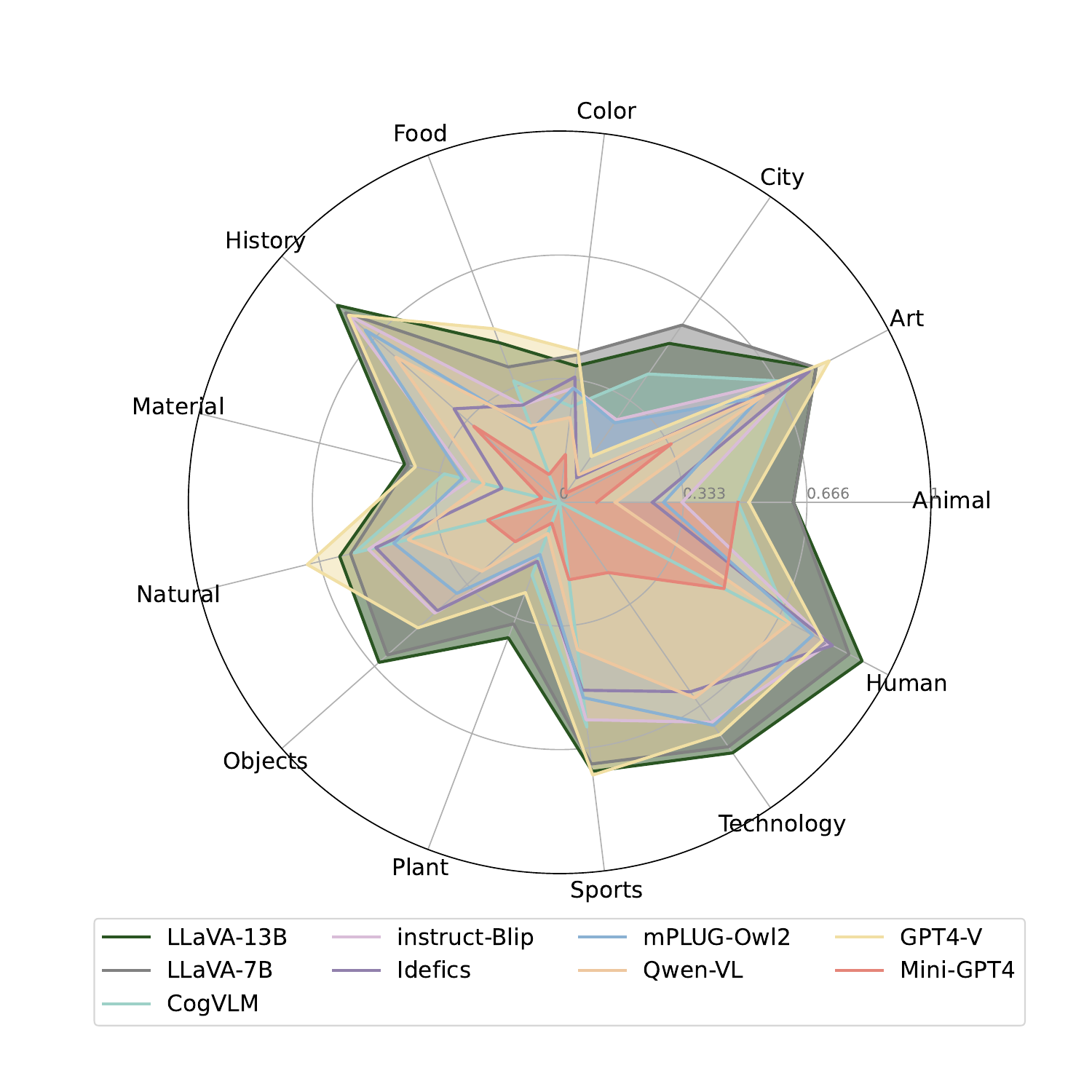}
    \vspace{-8mm}
    \caption{Accuracy of MLLMs on natural spurious images in our proposed benchmark \benchname. The higher accuracy indicates that MLLMs answer correctly when accompanied by spurious images.}
    \label{rader}
    \vspace{-6mm}
\end{figure}

Despite the success achieved by state-of-the-art MLLMs, most studies mainly focus on simple VQA. However, MLLMs are usually applied to complex vision reasoning scenarios, where the answers are usually not included in the images, which requires MLLMs to utilize the reasoning ability of LLM to answer.
We identify an visual illusion, the instinctive bias, which is widespread in vision reasoning. Existing MLLMs are prone to ignore the semantic information in reasoning quizzes and answer directly to the objects in the pictures instead of utilizing their reasoning ability.  In \autoref{examplecase}, we show a specific example of instinctive bias. Under the text-only condition, LLaVA can accurately answer the correct answer (\textit{i.e.} Cairns). However, when the image only contains the spurious image, LLaVA assumes Eiffel tower to be the corresponding answer and ignores the semantics of the question. 
This type of illusion affects the widespread use of MLLMs. In scenarios such as shopping recommendations and real-time VQA, users want to be recommended similar styles of schoolbags, or users cannot describe accurately and choose to upload pictures for information supplementation. With the instinctive bias, MLLMs tend to give incorrect answers. Therefore, it is essential to establish a benchmark to quantify the impact of such issues in current MLLMs.



To study the illusion of MLLMs under spurious visual inputs, we design a novel benchmark called \benchname{}. \benchname{} collects over 7,000 question-answer (QA) pairs in 13 categories, where each pair contains multiple answer-related images that may mislead MLLMs. We first use GPT-4~\citep{2303.08774} to generate meaningful QA pairs with five related but incorrect answers and a correct one. Based on the generated answers, we leverage the advanced diffusion model to generate the corresponding spurious images for each question. Specifically, we generate factual images with the correct answers as a comparison.
In addition to natural images, we also generate five typographic images for spurious answers, inspired by \citet{mmsafety}. To ensure that the synthetic data is not biased, we collect corresponding realistic images from the Internet via search engine.
Based on the design benchmark, we conducted an in-depth analysis to uncover the instinctive bias present in mainstream MLLMs. 
Our findings, presented in \autoref{rader}, demonstrate that 9 state-of-the-art MLLMs including GPT-4V suffer from visual illusion when presented with spurious visual inputs. 
This phenomenon indicates that by providing information related to spurious answers, images can induce MLLMs to instinctively focus on the visual content, resulting in responses that are predominantly based on visual information without proper reasoning and thinking. This is similar to the cases of unconscious decision-making processes observed in human brains~\citep{kahneman2011thinking,booch2021thinking}.

Our contributions are summarized as follows:
\textbf{1)} We first identify the visual instinctive bias in MLLMs, where spurious visual inputs can cause current MLLMs to delude.
\textbf{2)} 
We propose \benchname{} to quantify the seriousness of instinctive bias across different types, demonstrating that this issue is universal across MLLMs.
\textbf{3)} We provide an in-depth analysis of the recent 9 representative MLLMs on our benchmark, showing their susceptibility to spurious visual inputs under different scenarios.

\begin{figure*}[t]
    \centering
    \includegraphics[width=\textwidth]{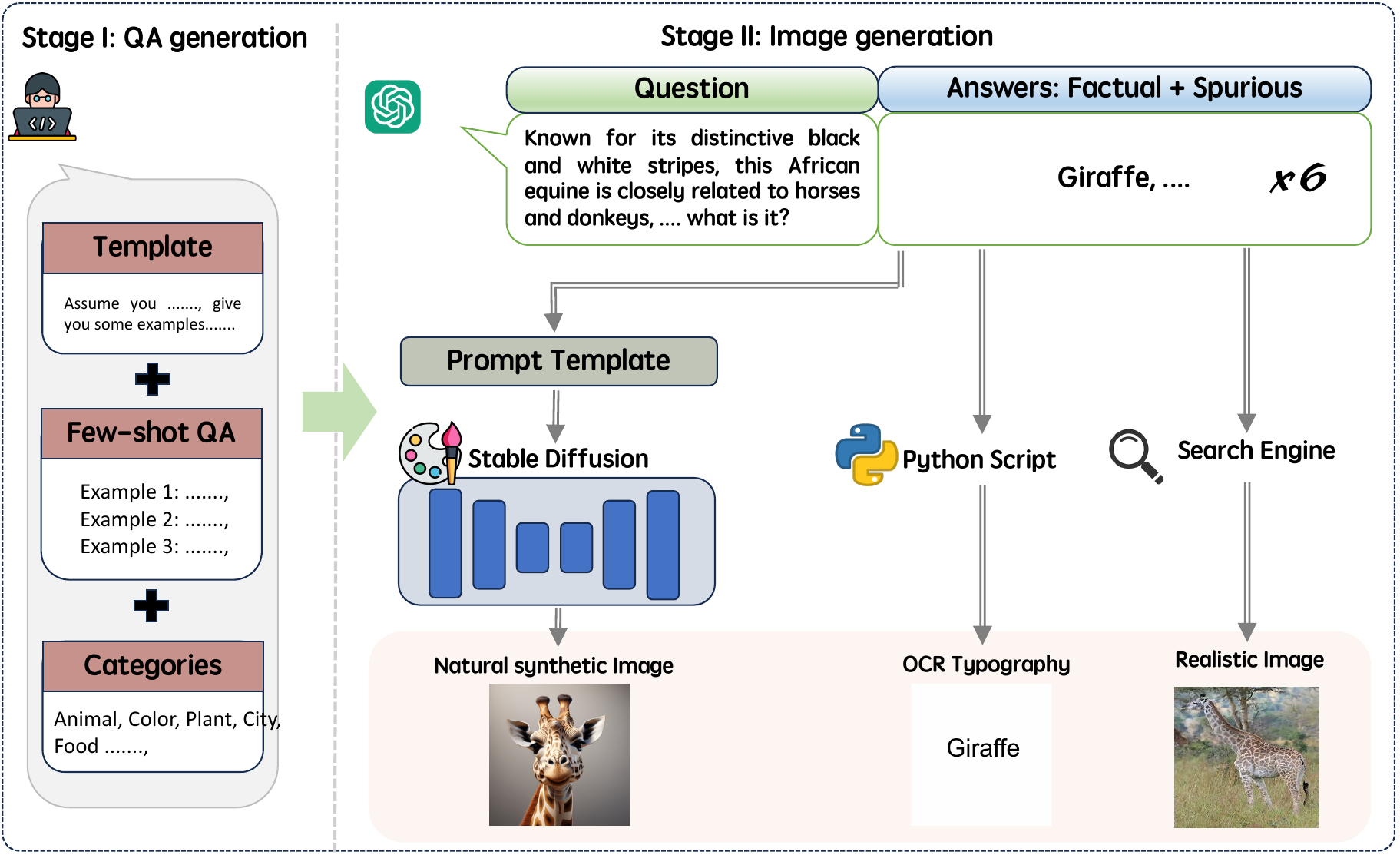}
    \vspace{-5mm}
    \caption{Pipeline of our dataset construction. First, we utilize GPT-4 to generate a set of QA pairs with five spurious answers. Next, we leverage image generators to generate corresponding images based on these answers (natural synthetic and typography). We use the answers as the keywords to obtain realistic images from search engine. Using these images, we construct a set of text and image pairs to evaluate the robustness of MLLMs to spurious images.}
    \vspace{-5mm}
    \label{framework}
\end{figure*}


\section{Method}
\label{section3}
In this section, we first present the background of multi-modal large language models (MLLMs) in commonsense question-answering (CQA) and the motivation of our study (\ref{moti}). Next, we introduce the proposed automated pipeline to generate our \benchname{} benchmark (\ref{process}).
Finally, we provide the designed evaluation metrics to measure the sensitivity of MLLMs on spurious images (\ref{metric}).

\subsection{Motivation} \label{moti}
By projecting the visual tokens into language space, existing MLLMs are able to equip large language models with visual processing ability.
However, past studies only demonstrate their ``fast thinking'' abilities in simple CQA tasks, but have yet explored their ``slow reasoning'' performance in complicated visual questions-answer tasks, such as when the input image provides relevant but indirect information about the correct answer.

Our study is motivated by the observation that current MLLMs, such as GPT-4V~\cite{2303.08774} and LLaVA~\cite{llava15}, are prone to inaccurate when presented with answer-correlated but answer-contradicted images. Examples depicted in \autoref{examplecase} demonstrate that LLaVA would fail spectacularly given a query accompanied by a spurious image. On the other hand, it is able to give correct answers in text-only scenarios. This indicates that the injection of additional image information has a detrimental effect on the capabilities of MLLMs.

To further study the role of the input image, we split the images into the following three types:
\textbf{1) Factual image}: the images are relevant and directly correspond to the correct answer,
\textbf{2) Spurious image}: the images are related to the question but do not correspond to the correct answer,
and \textbf{3) Random image}: the images are unrelated to either the question or answer. 

We then construct a set of image-text pairs to evaluate the performance of MLLMs under these three kinds of scenarios.

\subsection{\benchname} \label{process}
In order to obtain a large dataset of image-text pairs, we have designed a three-step automatic pipeline for generating and collecting the necessary data. The overall pipeline is shown in \autoref{framework}. We first pre-define 13 meta-categories for the proposed dataset, where the distribution of each category is illustrated in \autoref{dataset}. As we notice MLLMs favor spurious answers that occurred in the images over the semantics of questions, we firstly generate CQA pairs which can be prompted directly to LLMs. Secondly, for each question, we generate 5 images corresponding to five wrong answers and one image corresponding to the correct answer as visual inputs for MLLMs. Additionally, we collect realistic images of the wrong answer for each question.

\paragraph{Step1: Text Pairs} \label{step1}
To fully utilize the superior language comprehension capabilities of GPT-4, we employ this state-of-the-art language model to assist in data creation. Specifically, we use it to generate around 100 unique question-answer (QA) pairs for each scenario given some QA pair examples. These questions are demonstrated to be neither too simple nor stray from factual accuracy. Then, we also instruct GPT-4 to provide an accurate answer along with five spurious alternatives for each question, serving as the primary entities for subsequent image creation steps. 
The prompts and some examples are detailed in \autoref{appendix}.

\paragraph{Step2: Image Generation and Collection}
Given the constructed QA-pairs, this step leverages the image generator to create corresponding images. We follow~\citet{mmsafety} to build two kinds of images: natural and typographies. Specifically, we apply the cutting-edge image generation model, Stable Diffusion (SD)~\cite{ldm} as the image generator. We integrate six answers obtained in the first step into a prompt template for SD. Then, we leverage SD to output images with a resolution of 1024x1024 for better detail restoration and later resize the images to 512x512 for storage. 

Additionally, We utilize a search engine to collect the realistic images corresponding to the answer from the Internet and resize the longest side to 512. We present some image-text pairs of CorrelatioQA in \autoref{datasetexample}

\paragraph{Step3: Typography Generation}
There are numerous scenarios such as road sign recognition and document scanning, where text within images plays a crucial role in practical applications. Additionally, testing with OCR images can better simulate complex real-world data environments, challenging the robustness of MLLMs. Therefore, we generate typography images.

Following ~\citet{mmsafety}, we use the Pillow library to print the answers on a plain white background like OCR images. The image size is set to 512x512, as detailed image refinement is not as critical in this step compared to the previous one. The font size is set to 90 to ensure text legibility and prominence in the images.

\subsection{Evaluation Metrics} \label{metric}
\paragraph{Successful Answer Rate} To analyze the assessment of \benchname, we employ successful answer rate as the metric to determine MLLMs' susceptibility to the instinctive bias, which is also referred to \textbf{Accuracy} defined as follows:
\begin{gather}
    Acc = \frac{C}{T},
\end{gather}
where $C$ denotes the number of image-text pairs correctly answered by the model, and $T$ represents the total number of image-text pairs. We further impose a word count limit for MLLMs' outputs as all labels in the benchmark do not exceed a length of five words. To count the number of $C$, we adopt an approximate match approach, where it is acceptable for the response to be an abbreviation of the label or any sentence containing the label. For instance, if the label is "Los Angeles Lakers" then responses such as "Lakers" or "It is Los Angeles Lakers" are both considered correct. 

\paragraph{Accuracy Drop}
To evaluate the sensitivity of MLLMs under spurious images, we further design an Accuracy Drop (AccDrop) metric as follows:
\begin{gather}
    AccDrop = A_{f} - A_{s},
\end{gather}
where $A_{f}$ and $A_{s}$ denote Accuracy on factual and spurious data respectively. A higher AccDrop value indicates superior model performance with factual data and poorer with spurious one, which reflects the sensitivity to deceptive type information.

\section{Experiments}

\subsection{Dataset Collection}

As outlined in \autoref{section3}, our approach involves several steps. First, we pre-collect a set of demonstrating question-answer (QA) pairs. We then use these pairs to guide GPT-4 in generating additional QA pairs across different categories, each with one correct answer and five incorrect answers. Based on the generated answers, we utilize a state-of-the-art Stable Diffusion model and OCR-generated script to generate corresponding factual and spurious images, respectively. For further details on the collected scenario and dataset statistics, please refer to \autoref{dataset}.

\subsection{Experimental Setup}
\paragraph{Models.}

We perform a comprehensive evaluation of 9 recently released MLLMs on our \benchname, including LLaVA-1.5-7B and 13B~\cite{llava15} (referred as LLaVA-7B and LLaVA-13B for convenience), MiniGPT-4~\cite{zhu2023minigpt},  mPLUG-Owl2~\cite{mplug}, Qwen-VL\cite{qwen}, Idefics~\cite{idefics}, GPT-4V~\cite{2303.08774}, InstructBlip ~\cite{instructionblip} and CogVLM~\cite{wang2023cogvlm}.

\paragraph{Parameter Settings.}

Considering the different versions and updates of MLLMs, we choose their latest released weights for testing. All other parameters for each model are set to default values as specified by the original authors. For the open-sourced model, if not specifically mentioned, we adopt the widely-used 7B version of LLM for evaluation.


Regarding image generation, \textbf{playground-v2-1024px-aesthetic} checkpoint is adopted in Stable Diffusion. Compared to the commonly used \textbf{stable-diffusion-xl-base-1.0} checkpoint, this checkpoint enables more realistic image generation quality and avoids simple and counter-intuitive results.

\paragraph{Prompt Settings.}

For QA pairs generation, we utilize GPT-4 to generate thousands of QA pairs by providing several demonstrating examples.

\begin{mdframed}[backgroundcolor=gray!20]
Give you some examples of QA pairs. The content of QA pairs should include truth and commonsense. No repeated examples and answers. The description of the question should be complex as much as possible. Here are some examples: [Q: Sample Question1 A: Correct Answer1 ], [Q: Sample Question2 A: Correct Answer2], give 100 examples in the format: [Q:, A:, W:], while W means you should also give other 5 wrong confusing answers. Reference these to generate 100 similar examples relevant to [Categories], [Detailed Requirement].
\end{mdframed}

\vspace{12pt}

For image generation, we present the correct and spurious answers under the following prompt template to the diffusion model.
\begin{mdframed}[backgroundcolor=gray!20]
A photo of [Spurious Answer], detailed, 8k, realistic, trending on artstation.
\end{mdframed}

\vspace{12pt}
For visual question answering, we adopt the following prompt template with the questions as the text inputs into MLLMs.
\begin{mdframed}[backgroundcolor=gray!20]
[Question] Answer in no more than five words. 
\end{mdframed}

Each question is along with the generated natural or typography image. To more accurately assess the model responses, we require MLLMs to directly answer the questions. This approach is reasonable since all the correct are less than five words.

\begin{figure*}[!t]
    \centering
    \includegraphics[width=\textwidth]{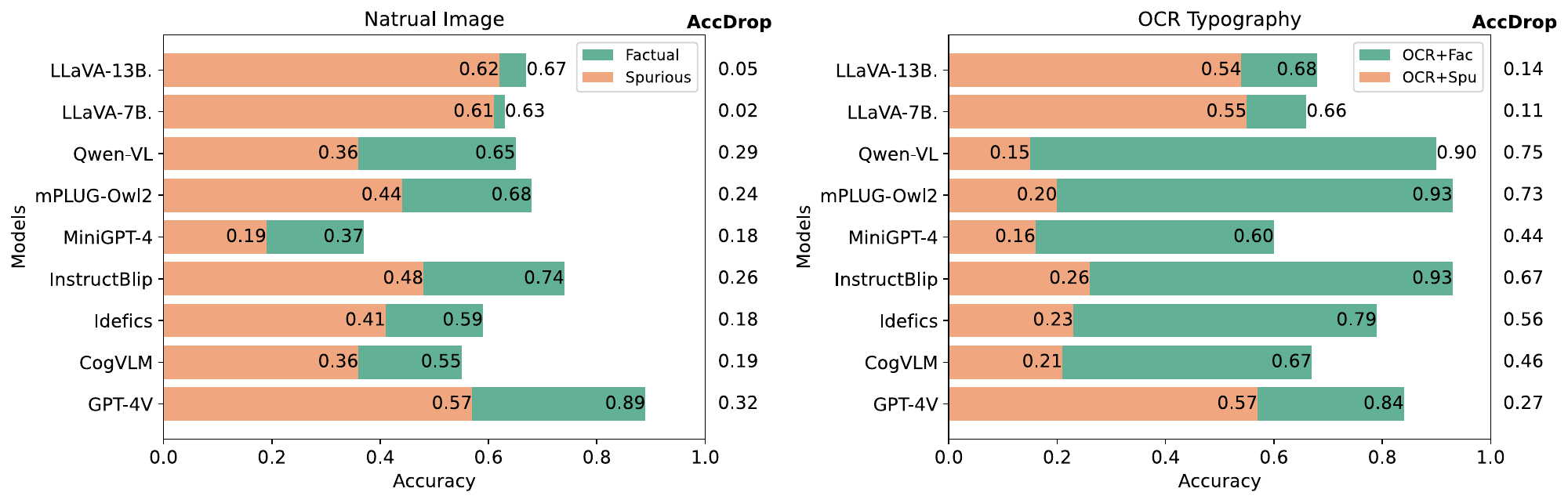}
    \vspace{-8mm}
    \caption{Assessments results on accuracy (Acc) and accuracy drop (AccDrop) for MLLMs. The results on the left refer to the natural image and the right one refers to the typography image. Corresponding AccDrop is presented on the right side of each figure. Fac and Spu denote factual and spurious, respectively.}
    \label{adi}
    \vspace{-4mm}
\end{figure*}

\subsection{Experimental Results}
\subsubsection*{Evaluation Results on \benchname}

In \autoref{adi}, we first present the overall accuracy (Acc) and accuracy drop (AccDrop) of nine MLLMs on our \benchname{}. The green color bars in each image represent the AccDrop from the factual image to spurious images, revealing that MLLMs consistently struggle with instinctive bias from spurious images, even for GPT-4V. This instinctive bias problem also occurs on the OCR data, which have higher AccDrop.



It is worth noticing that LLaVA and GPT-4V have higher average accuracy on the spurious images compared with other MLLM. What's more, both LLaVA-7B and LLaVA-13B exhibit almost no fluctuation in both spurious and factual contexts, which we believe can be attributed to its training data. To enhance the model's capabilities across various domains, researchers incorporate datasets like OK-VQA~\cite{okvqa} and A-OKVQA~\cite{AOKVQA} which require extensive knowledge to answer the question. Such training data enables LLaVA to reduce the influence of unessential images and leverage the inherent capabilities of LLMs for reasoning, thus leading to similar accuracy for LLaVA in both factual and spurious images. However, other tested MLLMs are mostly trained on image-answer-consistent data, therefore showing a performance drop between factual and spurious images. For GPT-4V, its pronounced proficiency in image-text understanding and language processing logically predicates a diminished propensity for instinctive bias. 


Compared to different types of image formats, typography exhibits a more serious instinctive bias problem over natural images. One potential reason is that spurious OCR typography might lead to a more simplistic and crude understanding of MLLMs. OCR images inherently contain limited information due to their simplistic textual content in our cases (e.g., a single word). Because MLLMs are found to possess a certain degree of OCR recognition capability, when MLLMs process information on these inputs, the proportion of spurious elements in the visual information is higher compared to that in generated images, which makes MLLMs suffer from more instinctive bias. Similarly, as the content of OCR typography is easier to understand,  MLLMs achieve higher accuracy when along with factual typography. 

\begin{table}[t]
\centering
\footnotesize
\begin{tabular}{M{0.3\columnwidth}|M{0.25\columnwidth}|M{0.25\columnwidth}}
\toprule[1pt]
Class      & Questions & Images \\ 
\midrule
Animal     &     105      &    630    \\
Art        &      105     &     630   \\
Color      &      99     &     594   \\
City       &        90   &    540    \\
Food       &       100    &     500   \\
History    &        104   &     624   \\
Human      &        105     &    630   \\
Material   &        90   &     540   \\
Natural    &        100   &     600   \\
Objects    &        105   &      630  \\
Plant      &        105   &     630   \\
Sports     &        95   &      570  \\
Technology &        105   &    630    \\ 
\midrule
Total      &       1,218    &    7,308    \\
\bottomrule[1pt]
\end{tabular}
\caption{The statistics distribution of \benchname.}
\vspace{-4mm}
\label{dataset}
\end{table}

\subsubsection*{Results on Different Categories}
\label{experiment3}
\autoref{dataset1} and \autoref{dataset2} present AccDrop of 9 MLLMs on each category in detail. The results indicate that MLLMs exhibit varying degrees of sensitivity to different categories. We observe that MLLMs on categories such as animals, colors, food, and plants suffer from larger AccDrop as highlighted. On the contrary, it shows significantly lower AccDrop in categories like history and art. Intuitively, the former categories consist of tangible entities while the latter include concepts like the ‘Industrial Revolution’ or 'The Lord of the Rings,' which may not be easily represented in generated natural images. 

Our analysis also shows that the impact of typography images on MLLMs is greater than that of natural data, where each category exhibits larger gap in AccDrop. Interestingly, unlike natural images, AccDrop in typography images does not show a significant difference across different categories. This is reasonable, as the content of typography images typically consists of words, which are easier to interpret compared to natural images. 

We hypothesize that the influences of the training data, cross-modal alignment training, and instruction tuning cause MLLMs to focus more on the semantic correlations between the query and the image. Identifying common patterns in the behavior of MLLMs could greatly assist in refining approaches for future work and is therefore an important finding.

\begin{figure}[t]
    \centering
    \includegraphics[width=\columnwidth]{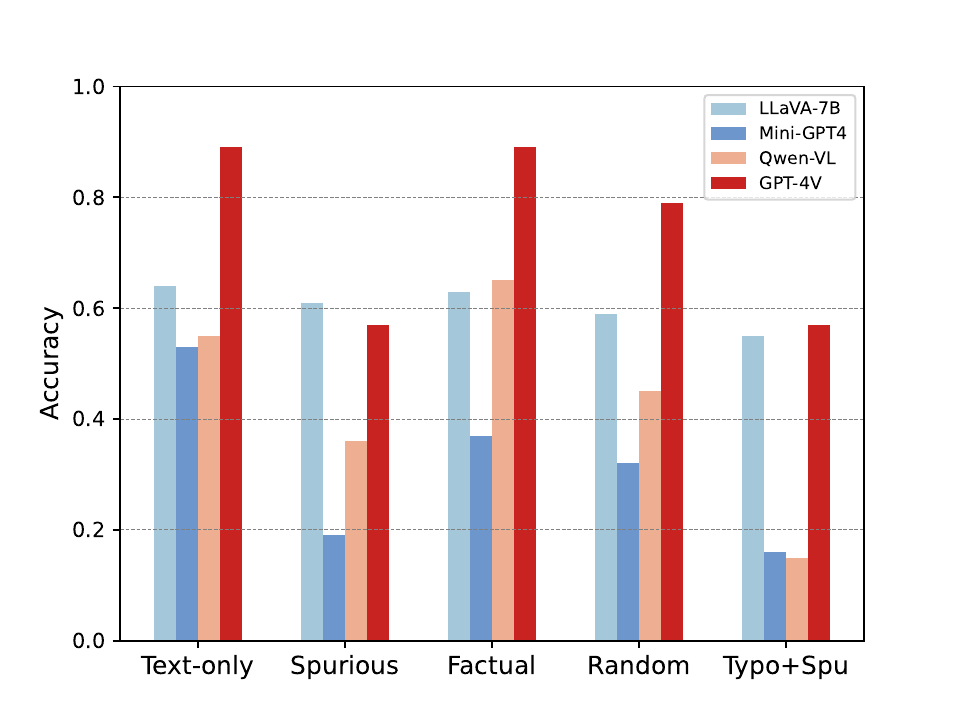}
    \vspace{-9mm}
    \caption{Accuracy of different input types. Typo+Spu indicates spurious OCR typography image.}
    \vspace{-6mm}
    \label{exp2}
\end{figure}

\begin{table*}[!ht]
\footnotesize
\centering
\resizebox{\textwidth}{!}{
\begin{tabular}{c|ccccccccccccc|c}
\toprule[1pt]
Image \textcolor{red}{$\downarrow$} & Animal & Art & Color & City & Food & History&Human & Material & Natural & Objects & Plant & Sports & Technology & Average \\
\midrule
CogVLM & \cellcolor{lightblue}0.39 & 0.06 & \cellcolor{lightblue}0.41 & 0.18 & 0.30 & $0^{*}$&0.17 & 0.32 & 0.08 & $0^{*}$ & \cellcolor{lightblue}\textbf{0.52} & 0.14 & $0^{*}$ & 0.19 \\
Idefics & \cellcolor{lightblue}\textbf{0.51} & 0.01 & \cellcolor{lightblue}0.04 & 0.12 & 0.33 & -0.01&0.05 & 0.25 & 0.13 & 0.26 & \cellcolor{lightblue}0.38 & 0.24 & 0.08 & 0.18 \\
InstructBlip & \cellcolor{lightblue}0.53 & 0.03 & \cellcolor{lightblue}0.53 & 0.15 & 0.34 & 0.07&0.03 & 0.36 & 0.10 & 0.36 & \cellcolor{lightblue}\textbf{0.54} & 0.23 & 0.07 & 0.26 \\
MiniGPT-4 & \cellcolor{lightblue}\textbf{0.45} & 0.09 & \cellcolor{lightblue}0.08 & 0.09 & 0.32 & 0.03&0.02 & 0.19 & 0 & 0.40 & \cellcolor{lightblue}0.36 & 0.32 & 0.07 & 0.18 \\
mPLUG-Owl2 & \cellcolor{lightblue}0.51 & 0.07 & \cellcolor{lightblue}\textbf{0.59} & 0.13 & 0.26 & 0.01&0.05 & 0.26 & 0.12 & 0.40 & \cellcolor{lightblue}0.42 & 0.26 & 0.05 & 0.24 \\
Qwen-VL & \cellcolor{lightblue}0.48 & 0.09 & \cellcolor{lightblue}0.43 & 0.21 & 0.46 & 0.02 &0.15& 0.45 & 0.16 & 0.39 & \cellcolor{lightblue}\textbf{0.56} & 0.39 & 0.04 & 0.29 \\
LLaVA-7B & \cellcolor{lightblue}0.02 & 0.02 & \cellcolor{lightblue}\textbf{0.09} & -0.01 & 0.01 & 0.01&0.02 & 0.05 & -0.01 & 0.02 & \cellcolor{lightblue}0.05 & 0.04 & 0 & 0.02 \\
LLaVA-13B & \cellcolor{lightblue}0.03 & 0 & \cellcolor{lightblue}\textbf{0.10} & 0.05 & 0.03 & 0 & -0.01 & 0.07 & 0.03 & 0.05 & \cellcolor{lightblue}0.08 & 0.06 & 0.02 & 0.05 \\
GPT-4V & \cellcolor{lightblue}0.41 & 0.14 & \cellcolor{lightblue}\textbf{0.74} & 0.15 & 0.36 &0.12& 0.11 & 0.40 & 0.17 & 0.39 & \cellcolor{lightblue}0.54 & 0.20 & 0.16 & 0.32\\
\midrule
Average & \cellcolor{lightblue}0.37 & 0.06 & \cellcolor{lightblue}0.33 & 0.12 & 0.27 & 0.03&0.07 & 0.26 & 0.09 & 0.25 & \cellcolor{lightblue}\textbf{0.38} & 0.21 & 0.05 & 0.19 \\
\bottomrule[1pt]
\end{tabular}
}
\vspace{-2mm}
\caption{Accuracy Drop (AccDrop) of MLLMs under 12 categories when applied natural images. AccDrop is the accuracy drop from the factual image into the spurious image. A higher value reflects a higher sensitivity to deceptive information. The three most sensitive categories are highlighted in blue background. \textbf{Bold} values are the top performance drop for each model. $0^{*}$ represents zero accuracy on both factual and spurious images.}
\vspace{-2mm}
\label{dataset1}
\end{table*}
\vspace{1.5cm}
\begin{table*}[!ht]
\footnotesize
\centering
\resizebox{\textwidth}{!}{
\begin{tabular}{c|ccccccccccccc|c}
\toprule[1pt]
Typography \textcolor{red}{$\downarrow$} & Animal & Art & Color & City & Food & History &Human & Material & Natural & Objects & Plant & Sports & Technology & Average \\
\midrule
CogVLM & 0.62 & 0.44 & 0.65 & \cellcolor{lightblue}0.85 & 0.58 & $0^{*}$ &$0^{*}$ & \cellcolor{lightblue}\textbf{0.90} & 0.45 & $0^{*}$ & \cellcolor{lightblue}0.79 & 0.43 & $0^{*}$ & 0.46 \\
Idefics & 0.68 & 0.44 & 0.20 & \cellcolor{lightblue}0.73 & 0.74 & 0.21 &0.59 & \cellcolor{lightblue}\textbf{0.80} & 0.51 & 0.54 & \cellcolor{lightblue}\textbf{0.80} & 0.48 & 0.56 & 0.56 \\
InstructBlip & 0.74 & 0.48 & 0.67 & \cellcolor{lightblue}0.73 & 0.75 & 0.35 &0.38& \cellcolor{lightblue}\textbf{0.83} & 0.56 & 0.75 & \cellcolor{lightblue}0.82 & 0.58 & 0.74 & 0.67 \\
MiniGPT-4. & 0.54 & 0.35 & 0.24 & \cellcolor{lightblue}0.54 & 0.54 & 0.23 &0.40& \cellcolor{lightblue}\textbf{0.66} & 0.43 & 0.35 & \cellcolor{lightblue}0.41 & 0.59 & 0.42 & 0.44 \\
mPLUG-Owl2 & 0.80 & 0.56 & 0.84 & \cellcolor{lightblue}\textbf{0.92} & 0.79 & 0.57 &0.65 & \cellcolor{lightblue}0.85 & 0.53 & 0.70 & \cellcolor{lightblue}0.84 & 0.65 & 0.72 & 0.73 \\
Qwen-VL & 0.79 & 0.65 & 0.91 & \cellcolor{lightblue}0.84 & 0.82 & 0.54 &0.78 & \cellcolor{lightblue}\textbf{0.95} & 0.65 & 0.74 & \cellcolor{lightblue}0.88 & 0.65 & 0.56 & 0.75 \\
LLaVA-7B & 0.12 & 0.12 & 0.11 & \cellcolor{lightblue}0.03 & -0.02 & 0.03&\textbf{0.48} & \cellcolor{lightblue}0.10 & 0.01 & 0.08 & \cellcolor{lightblue}0.21 & 0.04 & 0.03 & 0.11 \\
LLaVA-13B & 0.06 & 0.21 & 0.29 & \cellcolor{lightblue}0.16 & 0.04 & 0.11&\textbf{0.56} & \cellcolor{lightblue}0.05 & 0.09 & -0.05 & \cellcolor{lightblue}0.13 & 0.03 & 0.02 & 0.14 \\
GPT-4V & 0.10 & 0.33 & \textbf{0.70} & \cellcolor{lightblue}0.18 & 0.26 & 0.09 &0.10 & \cellcolor{lightblue}0.54 & 0.24 & 0.08 & \cellcolor{lightblue}0.19 & 0.16 & 0.37 & 0.27 \\
\midrule
Average & 0.43 & 0.36 & 0.46 & \cellcolor{lightblue}0.49 & 0.45 & 0.22 &0.44 & \cellcolor{lightblue}\textbf{0.57} & 0.35 & 0.32 & \cellcolor{lightblue}0.50 & 0.36 & 0.34 &0.46  \\

\bottomrule[1pt]
\end{tabular}
}
\vspace{-2mm}
\caption{Accuracy Drop (AccDrop) of MLLMs under 12 categories when applied typography. AccDrop is the accuracy drop from the factual image into the spurious image. A higher value reflects a higher sensitivity to deceptive information. The three most sensitive categories are highlighted in blue background. \textbf{Bold} values are the top performance drop for each model. $0^{*}$ represents zero accuracy on both factual and spurious images.}
\vspace{-3mm}
\label{dataset2}
\end{table*}

\subsubsection*{Spurious Information induces Visual Illusion}
The variation in performance among MLLMs also motivates us to analyze the impact of image type on model accuracy. In \autoref{exp2}, we present a comprehensive comparison of the average accuracy of four MLLMs under five different conditions. The "Text-only" condition indicates that only the text query is used to prompt the model. Regarding the multi-modality condition, we provide the factual, spurious, and random images, respectively. For the random image, we randomly select an image from another category for a specific question. Notably, only in scenarios with text-only and factual image inputs do MLLMs have comparable performances. It suggests that the strategy of using images as supplementary information does not positively influence the models' responses even if the answer is hidden in the visual inputs. Compared to the other four conditions, spurious data induce more instinctive bias in the selected four MLLMs, particularly evident with OCR typography. 


For the text-only scenario, we sample 20\% of the questions from each category to test GPT-4 due to its request rate limit. The results indicate that GPT-4 achieves remarkably high accuracy in text-only scenarios with almost all questions being correctly answered. GPT-4V, one of the most advanced MLLMs currently available, demonstrates a lower average accuracy than LLaVA when spurious images are added. This is noteworthy as larger models with superior language processing capabilities are generally expected to perform better, especially for those that are not specifically fine-tuned for particular tasks.


\subsection*{Results on Realistic Image}
In our main study, we utilized the Stable Diffusion model to synthesize a large number of images for studying the inductive bias problem in MLLMs. Additionally, to better align with MLLMs' real-world applications, we evaluated the accuracy and accuracy drop of MLLMs on realistic factual and spurious images.

Following the pipeline shown in \autoref{framework}, we first utilize GPT-4 to generate correct and incorrect text answers. Then, we employ a search engine to obtain corresponding realistic images using the search keywords from the correct and incorrect answers. Finally, we resize the images proportionally to ensure the shorter side remained at 512 pixels.

\autoref{realistic} shows the accuracy and accuracy drop for Qwen-VL and LLaVA-7B on both realistic and natural synthetic images. We observed that MLLMs exhibit similar behavior on both types of images, indicating that the conclusions drawn from the massive amount of synthetic images are generalizable to realistic images.
Furthermore, the performance drop between spurious realistic images and synthetic images may be due to the purity of the content in the searched realistic images. Images retrieved through keyword searches may contain information beyond the keywords themselves.
In \autoref{realcomp}, we provide some examples of real pictures and synthetic pictures under the same spurious answer.

\begin{table}[t]{
\footnotesize
\centering
\resizebox{0.5\textwidth}{!}{
\tabcolsep1pt
\begin{tabular}{c|cc|cc}
\toprule[1pt]
 \multirow{2}{*}{Image types} &  \multicolumn{2}{c|}{LLaVA-7B} &\multicolumn{2}{c}{Qwen-VL} \\
& Acc (Spu) &AccDrop \textcolor{red}{$\downarrow$} & Acc (Spu) & AccDrop \textcolor{red}{$\downarrow$} \\
\midrule
Realistic &0.55 & 0.06 & 0.33 & 0.34 \\
\addlinespace
Natural Synthetic &0.61 &0.02 &0.36 &0.29 \\
\bottomrule[1pt]
\end{tabular}
}}
\vspace{-2mm}
\caption{Acc on spurious images and AccDrop of LLaVA-7B and Qwen-VL on realistic images and natural synthetic images. ``Spu'' denotes spurious.}
\vspace{-4mm}
\label{realistic}
\end{table}

\begin{figure*}[t]
    \centering
    \includegraphics[width=\textwidth]{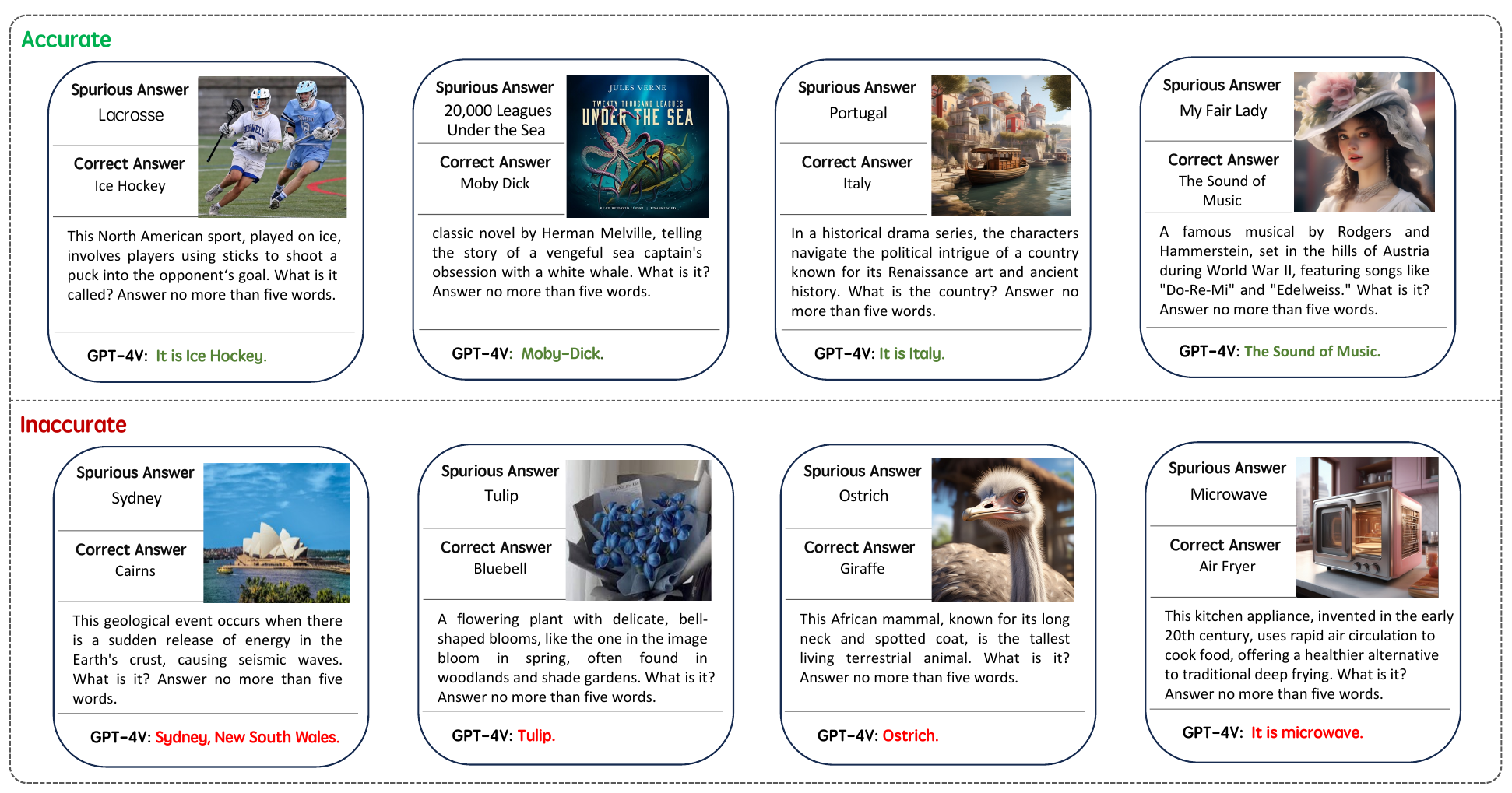}
    \vspace{-5mm}
    \caption{Visualization of image-text pairs in \benchname{}. The top row displays the examples where all tested MLLMs answer correctly, while the bottom row shows instances where MLLMs answer incorrectly. In each block, we provide the question, true label, spurious answer, the image generated by the spurious answer and responses of GPT-4V for each pair.}
    \label{exp3}
    \vspace{-5mm}
\end{figure*}

\subsubsection*{Qualitative Analysis}
\label{qualitative}
\autoref{exp3} further visualizes the examples where all 9 MLLMs answer correctly or incorrectly, respectively. For the image from accurate answers, we observe that the image contents do not significantly mislead the answers. For example, an image for “My Fair Lady” might be interpreted by MLLMs as “A woman wearing a medieval-style hat adorned with a flower,” leading to a shift in the relationship between the image and text towards “random” and “irrelevant” as we defined before.

In contrast, the images from the inaccurate examples are not only prominently recognizable but also discernible by the MLLMs' visual extraction modules. These findings briefly suggest that MLLMs are sensitive to images with tangible themes and prominent content, such as animals and objects. Categories like history and art, which are not as easily identifiable in images as physical objects, tend to have higher accuracy in responses.

\section{Related Work}
\paragraph{Multi-modal Large Language Models.} Benefiting from the exponential advancement of large language models (LLMs), 
a series of studies have introduced multi-modal large language models (MLLMs) by leveraging LLMs as their reasoning engine and textual interface~\cite{zhu2023minigpt,llava, wang2023visionllm, pi2023detgpt, pi2023perceptiongpt}.  

MLLMs achieve powerful visual understanding by training on image-text pairs. They can accurately extract semantic information from images and convert it into text that is easily comprehensible. Additionally, they utilize LLMs' reasoning ability to complete multi-modal tasks such as visual question-answering (VQA) and captioning.

\paragraph{Visual Illusion and Hallucination on MLLMs.} Some studies~\citep{yin2023survey, liu2024survey} demonstrate that MLLMs tend to provide responses that are inconsistent with visual information, which is known as illusion or hallucinations.
There are many works to study the MLLMs these problems. 
For example, \citet{pope} and \citet{gaive} propose benchmarks and introduce GPT-4V to detect and evaluate the responses for object hallucination. 
To alleviate the problem, \citet{pope} proposes an instruction fine-tuning strategy to balance the positive and negative samples in the training data. 
Contrary to these approaches, our work mainly concentrates on visual illusion when spurious visual inputs are presented.

\section{Conclusion}

In this paper, we demonstrate that current multi-modal large language models (MLLMs) are easy to raise instinctive bias through deceptive images. We first design an automatic pipeline that utilizes GPT-4 and Stable Diffusion to generate image-text pairs with factual and spurious images. Along with the designed pipeline, we construct a benchmark under 13 kinds of categories to evaluate the visual illusion of MLLMs under spurious visual inputs. Furthermore, we present a comprehensive analysis of the sensitivity to instinctive bias in MLLMs across various categories and under different conditions. 
We hope our work aids in better assessing the comprehensive capabilities of MLLMs in real-world scenarios and understanding the modality alignment of MLLMs.
Through our findings, future work could concentrate on adjusting training strategies, aiding MLLMs in appropriately calibrating their attention to image information based on its relevance in suitable contexts.

\section{Limitations}
Our research introduces the widespread instinctive bias in multi-modal large language models (MLLMs) towards deceptive images. We suggest that this may be associated with training data. However, MLLMs supporting other modalities such as video and audio may also exhibit instinctive bias due to their predominant use of data pairs with simple modality relationships in the training process, which is worth exploring in future work.
Additionally, our proposed \benchname, which consists of questions whose answers are entities, limits the evaluation to other types of questions. Due to the size of MLLMs, we do not conduct further assessments on larger parameter versions of large language models (i.e., Vicuna-33B). However, we do find that instinctive bias appears to be unrelated to the model scale (\autoref{adi}).
\section{Acknowledgements}
We express our gratitude to Lingyi Zhu, Yongjie Cai, Niya, Peiming Zhang and Han Peng from HKUST for verification and proofreading of the CorrelationQA benchmark.

\bibliography{anthology,custom}

\clearpage
\appendix

\begin{table*}[ht]{
\footnotesize
\centering
\resizebox{\textwidth}{!}{
\begin{tabular}{c|ccccccccccccc|c}
\toprule[1pt]
 Image& Animal & Art & Color & City & Food & History&Human & Material & Natural & Objects & Plant & Sports &Tech. & Average \\
\midrule
CogVLM & 0.48 & \cellcolor{lightblue}\textbf{0.70} & 0.42 & 0.26 & 0.35 & 0 &\cellcolor{lightblue}\textbf{0.70}& 0.32 & 0.57 & 0 & 0.21 & 0.61 & \cellcolor{lightblue}0 & 0.36 \\
Idefics & 0.25 &\cellcolor{lightblue} 0.76 & 0.08 & 0.34 & 0.28 & 0.38 &\cellcolor{lightblue}\textbf{0.83}& 0.16 & 0.51 & 0.44 & 0.17 & 0.51 & \cellcolor{lightblue}0.62 & 0.41 \\
InstructBlip & 0.33 &\cellcolor{lightblue} 0.73 & 0.27 & 0.31 & 0.28 &0.76&\cellcolor{lightblue}\textbf{0.82} & 0.25 & 0.53 & 0.45 & 0.15 & 0.59 & \cellcolor{lightblue}0.72 & 0.48 \\
MiniGPT-4 & 0.10 &\cellcolor{lightblue}0.34 & 0.03 & 0.13 & 0.08 & 0.31 &\cellcolor{lightblue}\textbf{0.50}& 0.05 & 0.20 & 0.16 & 0.06 & 0.21 &\cellcolor{lightblue} 0.23 & 0.19 \\
mPLUG-Owl2 & 0.28 & \cellcolor{lightblue}0.61 & 0.26 & 0.31 & 0.21 & 0.70 &\cellcolor{lightblue}\textbf{0.77}& 0.27 & 0.46 & 0.37 & 0.15 & 0.53 &\cellcolor{lightblue}0.73 & 0.44 \\
Qwen-VL & 0.15 &\cellcolor{lightblue} 0.62 & 0.09 & 0.23 & 0.22 &0.59 &\cellcolor{lightblue}\textbf{0.70}& 0.21 & 0.42 & 0.28 & 0.09 & 0.40 & \cellcolor{lightblue}0.64 & 0.36 \\
LLaVA-7B & 0.62 & \cellcolor{lightblue}0.78 & 0.58 & 0.40 & 0.39 & 0.77 &\cellcolor{lightblue}\textbf{0.88}& 0.42 & 0.58 & 0.62 & 0.35 & 0.71 &\cellcolor{lightblue} 0.80 & 0.61 \\
LLaVA-13B & 0.63 &\cellcolor{lightblue} 0.78 & 0.52 & 0.37 & 0.46 &0.80&\cellcolor{lightblue}\textbf{0.92} & 0.43 & 0.61 & 0.65 & 0.39 & 0.73 &\cellcolor{lightblue} 0.82 & 0.62 \\
GPT-4V & 0.51 & \cellcolor{lightblue}\textbf{0.82} & 0.15 & 0.41 & 0.50 &0.76 &\cellcolor{lightblue} 0.80& 0.40 & 0.70 & 0.51 & 0.26 & 0.74 &\cellcolor{lightblue} 0.76 & 0.57 \\
\midrule
Average &0.37&\cellcolor{lightblue}0.68&0.27&0.31&0.31&0.56&\cellcolor{lightblue}\textbf{0.77}&0.28&0.51&0.39&0.20&0.56&\cellcolor{lightblue}0.59& 0.45\\

\bottomrule[1pt]
\end{tabular}
}}
\caption{Accuracy (Acc) of MLLMs on \benchname{} under twelve categories when applied spurious image. We highlight the top three accuracy categories in blue background. \textbf{Bold} values are the maximum accuracy for each model.  }
\label{dataset3}
\end{table*}

\vspace{1.5cm}

\begin{table*}[!h]{
\footnotesize
\centering
\resizebox{\textwidth}{!}{
\begin{tabular}{c|ccccccccccccc|c}
\toprule[1pt]
Typography & Animal &Art & Color & City & Food & History &Human& Material & Natural & Objects & Plant & Sports & Tech. & Average \\
\midrule
CogVLM & 0.34 &\cellcolor{lightblue} \textbf{0.53} & 0.32 & 0.06 & 0.29 & \cellcolor{lightblue}0 &0 & 0.09 & 0.46 & 0 & 0.11 &\cellcolor{lightblue}\textbf{ 0.53} & 0 & 0.21 \\
Idefics & 0.19 &\cellcolor{lightblue} \textbf{0.48}& 0.03 & 0.03 & 0.14 & \cellcolor{lightblue}0.41&0.40 & 0.04 & 0.33 & 0.36 & 0.09 & \cellcolor{lightblue}0.38 & 0.17 & 0.23 \\
InstructBlip & 0.23 &\cellcolor{lightblue} 0.32 & 0.28 & 0.04 & 0.15 & \cellcolor{lightblue}0.48 &\textbf{0.56} & 0.10 & 0.35 & 0.22 & 0.08 & \cellcolor{lightblue}0.40 & 0.22 & 0.26 \\
MiniGPT-4 & 0.18 & \cellcolor{lightblue}0.34 & 0.09 & 0.03 & 0.05 & \cellcolor{lightblue}0.26 &\textbf{0.44} & 0.02 & 0.17 & 0.20 & 0.06 & \cellcolor{lightblue}0.14 & 0.18 & 0.16 \\
mPLUG-Owl2 & 0.18 & \cellcolor{lightblue}0.30 & 0.12 & 0 & 0.09 &\cellcolor{lightblue} \textbf{0.37}&0.31 & 0.08 & 0.27 & 0.22 & 0.05 & \cellcolor{lightblue}0.30 & 0.25 & 0.20 \\
Qwen-VL & 0.12 & \cellcolor{lightblue}0.25 & 0.07 & 0.03 & 0.13 &\cellcolor{lightblue} 0.22 &0.19 & 0.04 & 0.24 & 0.14 & 0.03 &\cellcolor{lightblue} \textbf{0.27 }& 0.25 & 0.15 \\
LLaVA-7B & 0.54 &\cellcolor{lightblue} 0.73 & 0.59 & 0.38 & 0.38 &\cellcolor{lightblue} 0.76 &0.49 & 0.42 & 0.59 & 0.57 & 0.26 &\cellcolor{lightblue}0.73 & \textbf{0.78} & 0.55 \\
LLaVA-13B & 0.60 &\cellcolor{lightblue} 0.56 & 0.43 & 0.32 & 0.47 &\cellcolor{lightblue} 0.71 &0.37 & 0.53 & 0.61 & 0.67 & 0.40 &\cellcolor{lightblue} 0.75 & \textbf{0.80} & 0.54 \\
GPT-4V & 0.70 &\cellcolor{lightblue} 0.57 & 0.19 & 0.39 & 0.58 &\cellcolor{lightblue} \textbf{0.86}&0.84 & 0.38 & 0.67 & 0.72 & 0.39 &\cellcolor{lightblue} 0.79 & 0.36 & 0.57 \\
\midrule
Average &0.34&\cellcolor{lightblue}0.45&0.23&0.14&0.27&\cellcolor{lightblue}0.45 &0.40&0.19&0.41&0.35&0.17&\cellcolor{lightblue}\textbf{0.48}&0.33&0.32 \\
\bottomrule[1pt]
\end{tabular}
}}
\caption{Accuracy (Acc) of MLLMs on \benchname{} under twelve categories when applied spurious typography. We highlight the top three accuracy categories in blue background. \textbf{Bold} values are the maximum accuracy for each model.}
\label{dataset4}
\end{table*}

\begin{table*}[ht]
\footnotesize
\centering
\resizebox{\textwidth}{!}{
\begin{tabular}{c|ccccccccccccc|c}
\toprule[1pt]

Image \textcolor{red}{$\downarrow$} & Animal & Art &Color  & City & Food & History &Human& Material & Natural & Objects & Plant & Sports & Tech. & Average \\
\midrule
CogVLM&\cellcolor{lightblue}45\%&8\%&\cellcolor{lightblue}49\%&41\%&46\%&0\%&20\%&50\%&12\%&0\%&\cellcolor{lightblue}\textbf{71\%}&19\%&0\%&35\%\\
Idefics&\cellcolor{lightblue}67\%&1\%&\cellcolor{lightblue}33\%&26\%&54\%&-3\%&6\%&61\%&20\%&37\%&\cellcolor{lightblue}\textbf{69\%}&32\%&11\%&31\%\\
InstructBlip&\cellcolor{lightblue}62\%&4\%&\cellcolor{lightblue}66\%&33\%&55\%&8\%&4\%&59\%&16\%&44\%&\cellcolor{lightblue}\textbf{78\%}&28\%&9\%&35\%\\
MiniGPT-4&\cellcolor{lightblue}82\%&21\%&\cellcolor{lightblue}73\%&41\%&80\%&9\%&4\%&79\%&0\%&71\%&\cellcolor{lightblue}\textbf{86\%}&60\%&23\%&49\%\\
mPLUG-Owl2&\cellcolor{lightblue}65\%&10\%&\cellcolor{lightblue}69\%&30\%&55\%&1\%&6\%&49\%&21\%&52\%&\cellcolor{lightblue}\textbf{74\%}&33\%&6\%&35\%\\
Qwen-VL&\cellcolor{lightblue}76\%&13\%&\cellcolor{lightblue}83\%&48\%&68\%&3\%&18\%&68\%&28\%&57\%&\cellcolor{lightblue}\textbf{86\%}&49\%&6\%&45\%\\
LLaVA-7B&\cellcolor{lightblue}3\%&2\%&\cellcolor{lightblue}\textbf{13\%}&-3\%&2\%&1\%&2\%&11\%&-2\%&3\%&\cellcolor{lightblue}13\%&5\%&0\%&3\%\\
LLaVA-13B&\cellcolor{lightblue}5\%&0\%&\cellcolor{lightblue}16\%&12\%&6\%&0\%&-1\%&14\%&5\%&7\%&\cellcolor{lightblue}\textbf{17\%}&8\%&2\%&7\%\\
GPT-4V&\cellcolor{lightblue}45\%&15\%&\cellcolor{lightblue}\textbf{83\%}&27\%&42\%&14\%&12\%&50\%&20\%&43\%&\cellcolor{lightblue}68\%&21\%&17\%&36\%\\
\midrule
Average &\cellcolor{lightblue}50\%&8\%&\cellcolor{lightblue}53\%&28\%&45\%&3\%&7\%&49\%&13\%&35\%&\cellcolor{lightblue}\textbf{62}\%&28\%&8\%&30\% \\

\bottomrule[1pt]
\end{tabular}
}
\caption{Accuracy declined ratio (the ratio between AccDrop (AccDrop) and Accuracy (Acc) on factual image) in natural image. It reflects the proportion of accuracy decline when models are exposed to spurious image compared to factual ones. We highlight the top three accuracy categories in blue background. \textbf{Bold} values are the maximum AccDrop proportion for each model.}
\label{dataset5}
\end{table*}

\begin{table*}[!h]
\footnotesize
\centering
\resizebox{\textwidth}{!}{
\begin{tabular}{c|ccccccccccccc|c}
\toprule[1pt]

Typography \textcolor{red}{$\downarrow$} & Animal & Art & Color & City & Food &History &Human& Material & Natural & Objects & Plant & Sports & Tech. & Average \\
\midrule
CogVLM&65\%&45\%&67\%&\cellcolor{lightblue}\textbf{93\%}&67\%&0\%&0\%&\cellcolor{lightblue}91\%&49\%&0\%&\cellcolor{lightblue}88\%&45\%&0\%&69\%\\
Idefics&78\%&48\%&87\%&\cellcolor{lightblue}\textbf{96\%}&84\%&34\%&60\%&\cellcolor{lightblue}95\%&61\%&60\%&\cellcolor{lightblue}90\%&56\%&77\%&71\%\\
InstructBlip&76\%&60\%&71\%&\cellcolor{lightblue}\textbf{95\%}&83\%&42\%&40\%&\cellcolor{lightblue}89\%&62\%&77\%&\cellcolor{lightblue}91\%&59\%&77\%&72\%\\
MiniGPT-4&75\%&51\%&73\%&\cellcolor{lightblue}95\%&92\%&47\%&48\%&\cellcolor{lightblue}\textbf{97\%}&72\%&64\%&\cellcolor{lightblue}87\%&81\%&70\%&73\%\\
mPLUG-Owl2&82\%&65\%&88\%&\cellcolor{lightblue}\textbf{100\%}&90\%&61\%&68\%&\cellcolor{lightblue}91\%&66\%&76\%&\cellcolor{lightblue}94\%&68\%&74\%&78\%\\
Qwen-VL&87\%&72\%&93\%&\cellcolor{lightblue}97\%&86\%&71\%&80\%&\cellcolor{lightblue}96\%&73\%&84\%&\cellcolor{lightblue}\textbf{97\%}&71\%&69\%&83\%\\
LLaVA-7B&18\%&14\%&16\%&\cellcolor{lightblue}7\%&-6\%&4\%&\textbf{49\%}&\cellcolor{lightblue}19\%&2\%&12\%&\cellcolor{lightblue}45\%&5\%&4\%&17\%\\
LLaVA-13B&9\%&27\%&40\%&\cellcolor{lightblue}33\%&8\%&13\%&\textbf{60\%}&\cellcolor{lightblue}9\%&13\%&-8\%&\cellcolor{lightblue}25\%&4\%&2\%&21\%\\
GPT-4V&13\%&37\%&\textbf{79\%}&\cellcolor{lightblue}32\%&31\%&9\%&11\%&\cellcolor{lightblue}59\%&26\%&10\%&\cellcolor{lightblue}33\%&17\%&51\%&32\%\\
\midrule
Average&55\%&46\%&68\%&\cellcolor{lightblue}\textbf{72\%}&59\%&31\%&46\%&\cellcolor{lightblue}71\%&47\%&41\%&\cellcolor{lightblue}\textbf{72\%}&45\%&47\%&57\% \\

\bottomrule[1pt]
\end{tabular}
}
\caption{Accuracy declined ratio (the ratio between AccDrop (AccDrop) and Accuracy (Acc) on factual image) in typography. It reflects the proportion of accuracy decline when models are exposed to spurious image compared to factual ones. We highlight the top three accuracy categories in blue background. \textbf{Bold} values are the maximum AccDrop proportion for each model.}
\label{dataset6}
\end{table*}

\label{sec:appendix}
\begin{figure*}
    \centering
    \includegraphics[width=0.8\textwidth]{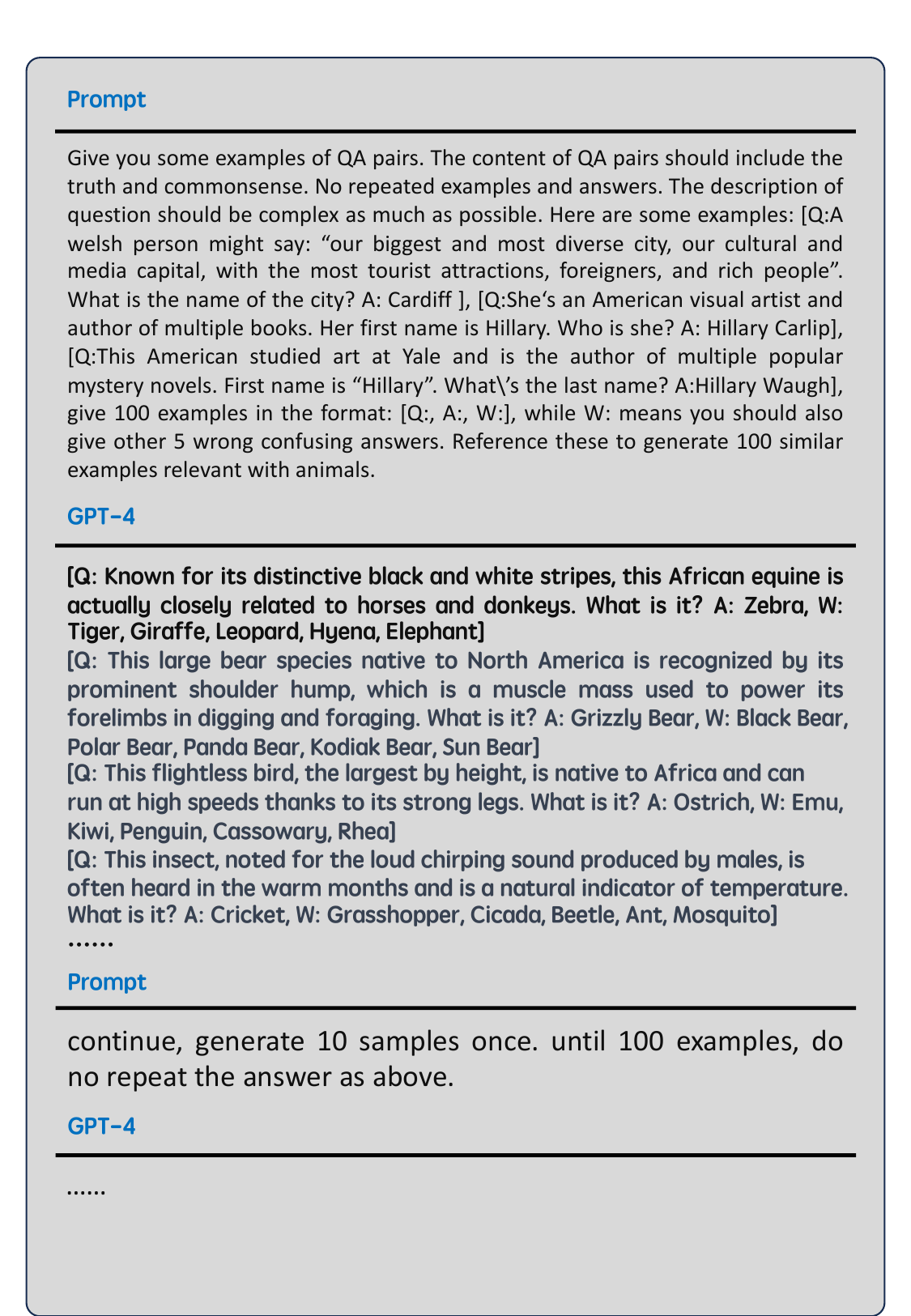}
    \caption{Prompt template and examples of QA pairs creation using GPT-4.}
    \label{appendix}
\end{figure*}

\section{More Related Works}
\paragraph{Adversarial Attack on LLMs.} Adversarial attacks are inputs that trigger the model to output something undesired~\cite{zou2023universal} even when developers impose constraints on model behaviors during the alignment process for safety purpose, such as reinforcement learning from human feedback (RLHF). 
Existing studies have shown that LLMs are still easily attacked to generate irrelevant or inappropriate outputs through methods like adversarial prompts~\cite{carlini2023poisoning,jailbreak,bertattack} and token manipulation~\cite{morris2020textattack}. 
On top of that, to bypass safeguarding mechanisms, various attack mechanisms~\cite{defending,defense} have been proposed to counteract user-driven adversarial behavior in both LLMs and MLLMs aspects. 
For example, \citet{mmsafety, pi2024mllmprotector} discovered that incorporating relevant images can trigger an image jailbreak in MLLMs, enabling the model to produce harmful information beyond what is achievable in a text-only scenario.

\section{Detailed Prompts Example}

\autoref{appendix} displays an example of generating question-answer (QA) pairs with GPT-4.  We detail the system prompt for the animal category and provide three example QA pairs for GPT-4 as references. Due to the output token limit, GPT-4 could only produce 10 QA pairs once, so we require it to continue generating more examples.

\section{More Experiments}
\subsection{Manual Verification}
 We randomly sample 20\% of the QA pairs from each category and verify if the actual answers match the true answers provided by GPT-4. The authenticity rates for QA pairs in each category are displayed in \autoref{manual}. Most of the categories have higher than 90\% authenticity rates except the class city. The reason is that in the city category, there are some fictional cities from the movies and novels besides the real world, which results in naming conflict.

\subsection{Accuracy Results on Spurious Image}
\label{append2}
In \autoref{dataset3} and \autoref{dataset4}, we present the accuracy of 9 MLLMs on spurious natural image and spurious typography, respectively. 

For the spurious natural image, categories like art, human, technology and history reach higher accuracy, which aligns with our analysis in \autoref{qualitative}. For categories such as art, technology and history, the spurious answers are often non-visualizable concepts (e.g., 5G technology, the Battle of Waterloo), and in the human category, spurious image containing portraits or photographs are unrecognizable to MLLMs, thus failing to significantly deceive or mislead the models.

For the spurious typography, accuracy across all 13 categories is more uniform. Compared to natural image, the application of typography results in a lower average accuracy for each category. We assume that since the content of typography solely consists of OCR text which does not involve understanding the content and is independent of the category, MLLMs are more directly misled by spurious information.

\subsection{Accuracy Declined Ratio}
We additionally defined \textbf{Accuracy declined ratio} to identify which categories experience the largest proportion of accuracy decline in MLLMs. The definition of accuracy declined ratio is as follows:
\begin{gather}
    ADR=\frac{AccDrop}{Acc_{f}},
\end{gather}
where $AccDrop$ denotes the pre-defined Accuracy drop metric of MLLMs, and $Acc_{f}$ represents accuracy on factual image. A higher accuracy declined ratio indicates more severely affected by spurious information, which is similar to AccDrop but emphasizes the relative effects.

\autoref{dataset5} and \autoref{dataset6} display the accuracy declined ratio results for natural image and typography. 
Our findings are consistent with those in \autoref{append2}. For natural image, categories like animal, color, and plant which consist of tangible entities experience a higher accuracy decline ratio. With typography, the accuracy decline ratio for all categories exceeds 30\%. After applying spurious images, the decline ratio for typography in every category is higher than for natural images.

\begin{figure*}[!t]
    \centering
    \includegraphics[width=\textwidth]{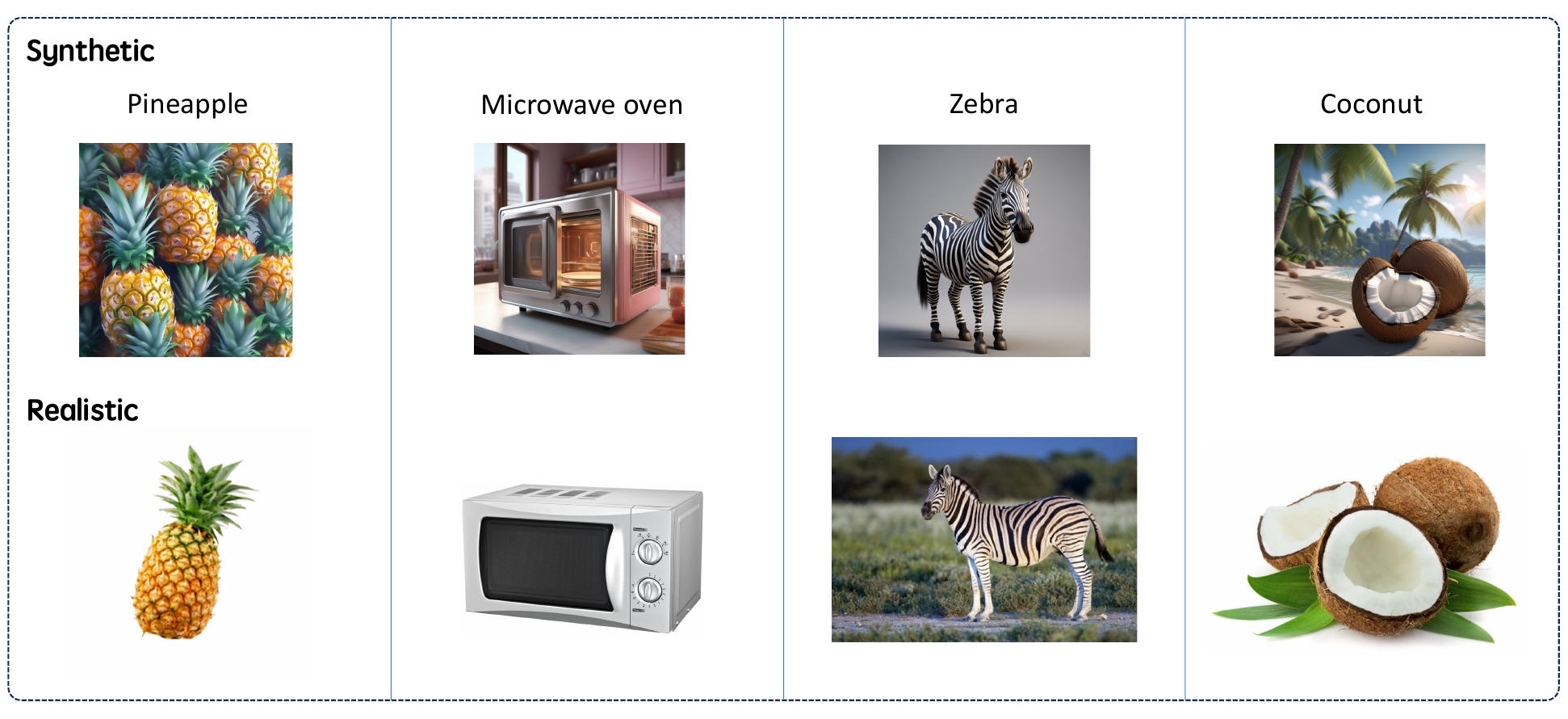}
    \vspace{-8mm}
    \caption{Examples of realistic pictures and synthetic pictures under the same spurious answer.}
    \label{realcomp}
    \vspace{-6mm}
\end{figure*}

\begin{table}[t]
\centering
\footnotesize
\begin{tabular}{M{0.3\columnwidth}|M{0.25\columnwidth}|M{0.25\columnwidth}}
\toprule[1pt]
Class      & Questions & Authenticity rate  \\ 
\midrule
Animal     &  105  & 100\%      \\
Art        &    105&  100\%     \\
City      &    90&  78\%      \\
Color       &   99&     95\%      \\
Food       & 100  &    95\%      \\
History    & 105   &    100\%     \\
Material   &  90  &    90\%     \\
Natural    &   100 &    100\%      \\
Objects    &   105 &    100\%   \\
Plant      &  105  &    91\%    \\
Sports     & 95  &     95\%     \\
Technology &   105 &    100\%       \\ 
\midrule
Average      &  101&     97\%     \\
\bottomrule[1pt]
\end{tabular}

\caption{We present the total number of questions and the Authenticity rate of \benchname. We randomly sample 20\% of QA pairs from each category and manually verify the Authenticity of true answers given by GPT-4. }
\label{manual}
\end{table}

\begin{figure*}[!t]
    \centering
    \includegraphics[width=\textwidth]{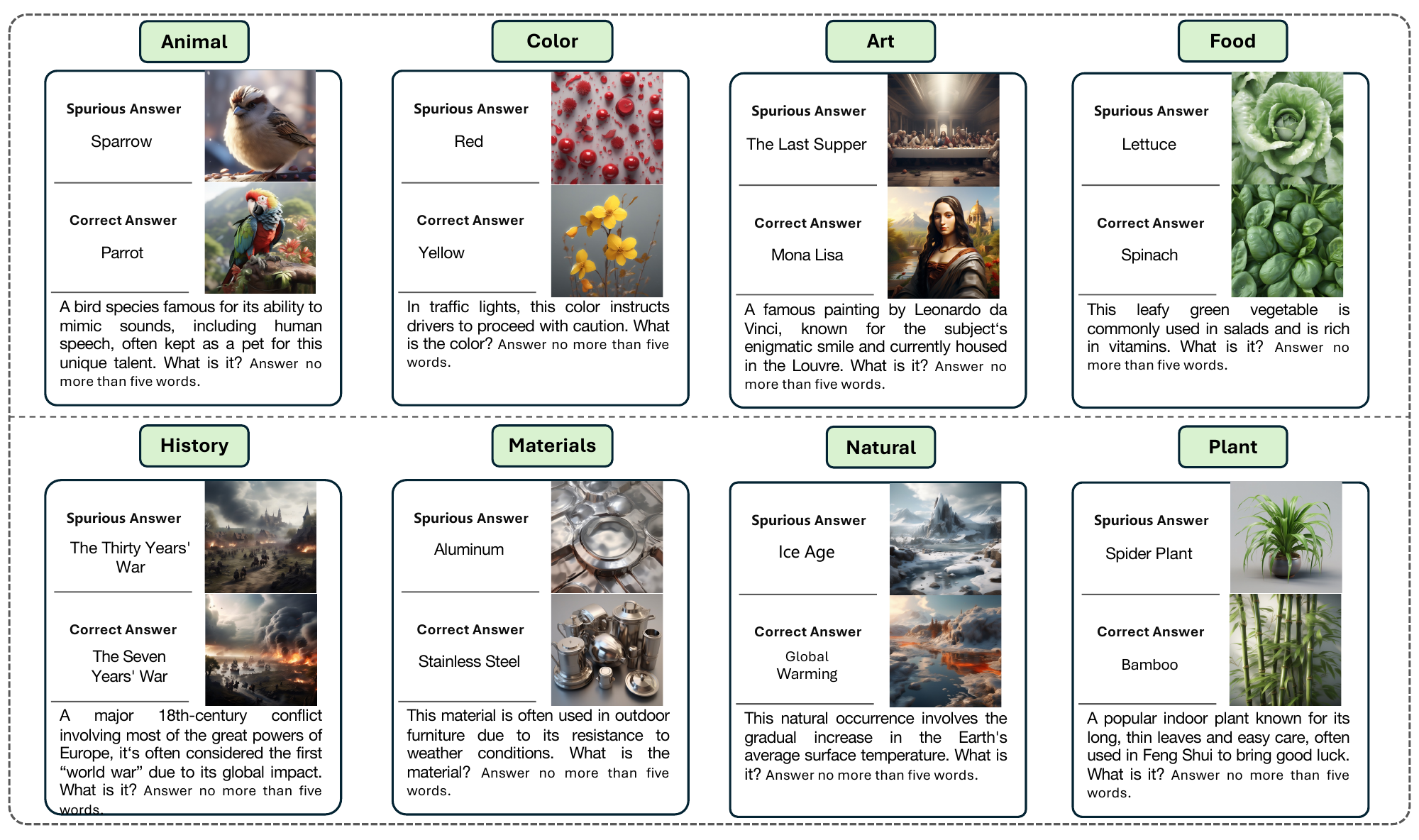}
    \vspace{-8mm}
    \caption{Examples of image-question pairs with the synthetic images of 8 categories.}
    \label{datasetexample}
    \vspace{-6mm}
\end{figure*}

\end{document}